\documentclass[11pt,a4paper]{article}
\usepackage[hyperref]{eacl2021}
\usepackage{times}
\usepackage{latexsym}

\usepackage{microtype}

\aclfinalcopy 

\usepackage{amsmath}
\usepackage{hyperref}
\usepackage{url}
\usepackage{mathrsfs}
\usepackage{multirow}
\usepackage{booktabs}
\usepackage{graphicx}
\usepackage{subcaption}
\usepackage{cleveref}
\usepackage{CJK}
\usepackage{CJKspace}
\usepackage{fdsymbol}
\usepackage{enumitem} 

\crefname{equation}{Eq.}{Eq.}
\crefname{section}{Section}{Sections}
\crefname{subsection}{Section}{Sections}
\crefname{subsubsection}{Section}{Sections}
\crefname{figure}{Figure}{Figures}
\crefname{table}{Table}{Tables}
\crefname{subfigure}{Figure}{Figures}
\crefname{algocf}{Algorithm}{Algorithms}
\usepackage{pifont}
\newcommand{\cmark}{\ding{51}}%
\newcommand{\xmark}{\ding{55}}%

\title{Cross-Lingual Vision-Language Navigation}

\author{An Yan
$^{\vardiamondsuit}$, Xin Eric Wang
$^{\spadesuit}$, Jiangtao Feng$^\heartsuit$, Lei Li$^\heartsuit$, William Yang Wang$^\clubsuit$\\
$^\vardiamondsuit$UC San Diego, $^\spadesuit$UC Santa Cruz, $^\clubsuit$UC Santa Barbara\\
\texttt{\small ayan@ucsd.edu, xwang366@ucsc.edu, william@cs.ucsb.edu}\\
$^\heartsuit$ByteDance AI Lab\\
\texttt{\small\{fengjiangtao,lileilab\}@bytedance.com}
}
\date{}

\begin{document}
\maketitle
\begin{abstract}
Commanding a robot to navigate with natural language instructions is a long-term goal for grounded language understanding and robotics.
But the dominant language is English, according to previous studies on vision-language navigation (VLN).
To go beyond English and serve people speaking different languages, we collect a cross-lingual Room-to-Room (XL-R2R) dataset, extending the original benchmark with new Chinese instructions.
Based on this newly introduced dataset, we study how an agent can be trained on existing English instructions but navigate effectively with another language under a zero-shot learning scenario.
Without any training data of the target language, our model shows competitive results even compared to a model with full access to the target language training data.
Moreover, we investigate the transferring ability of our model when given a certain amount of target language training data. \footnote{XL-R2R is released at \url{https://github.com/zzxslp/XL-VLN}}
\end{abstract}

\section{Introduction}
\begin{figure*}[t]
  \centering
  \includegraphics[width=.95\linewidth]{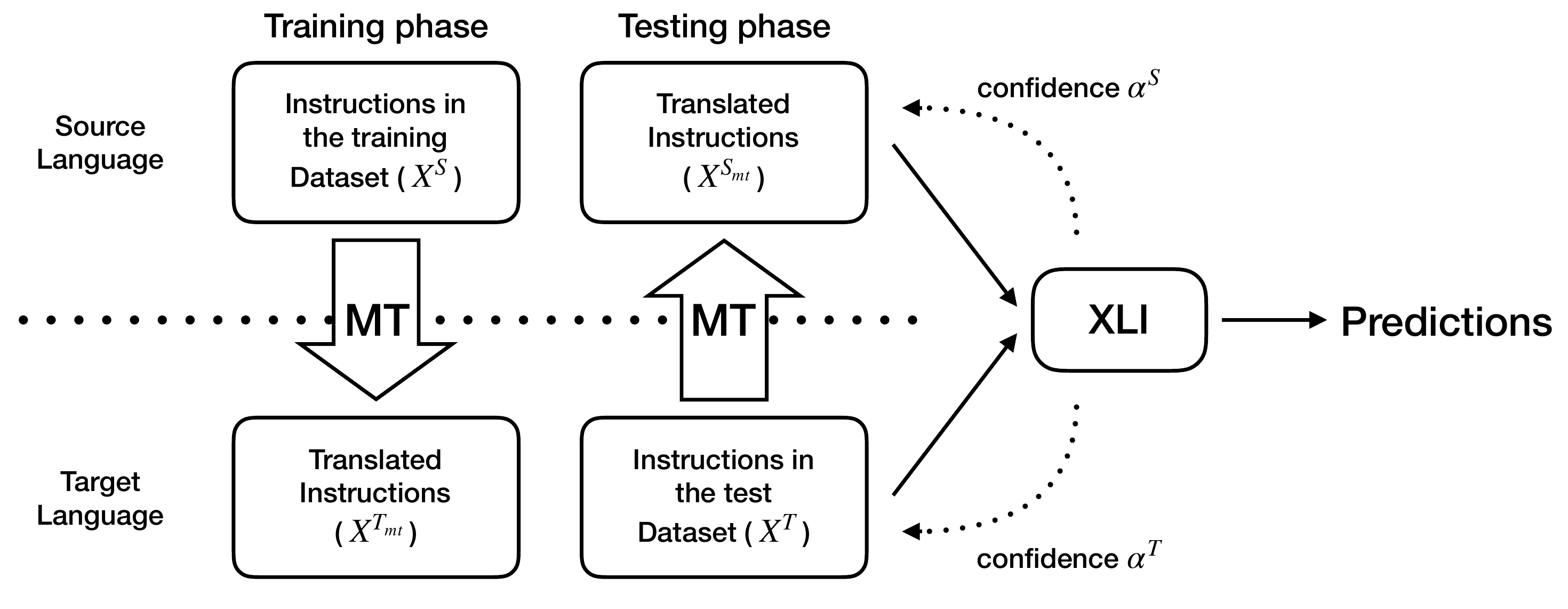}
  \caption{Overview of cross-lingual Instructor that learns to benefit from two learning schemes.}
  \label{fig:framework}
\end{figure*}

Grounded natural language understanding in the real world is an essential ability for a robot to communicate with humans~\citep{walkthewalk06,chen2011learning,artzi2013weakly}.
The task of vision-language navigation (VLN)~\citep{anderson2018vision}, which requires the agent to follow natural language instructions and navigate in houses, thrives recently due to photo-realistic simulation, free-form language instructions, and large-scale training.
The VLN task is particularly challenging and requires an understanding of both language instructions and visual dynamics as well as cross-modal alignment.  

Despite recent advances in VLN~\citep{fried2018speaker, wang2019reinforced, tan2019learning}, existing VLN benchmarks~\citep{anderson2018vision, chen2019touchdown} are all monolingual in that they only contain English instructions. The navigation agents are trained and tested with only English corpus and thus unable to serve non-English speakers.
To fill this gap, one can collect corresponding instructions in the language that the agent is expected to execute. But it is not scalable and practical as there are thousands of languages in the world, and collecting large-scale data for each language would be very expensive and time-consuming.

Therefore, in this paper, we introduce the task of cross-lingual VLN to endow an agent the ability to execute instructions in different languages.
First, \emph{can we learn an agent that is trained on existing English instructions but still able to navigate reasonably well for a different language?} 
This is essentially a zero-shot learning scenario where no training data of target language is available.

An intuitive approach is to train the agent with English data, and at test time, use a machine translation system to translate the target language instructions to English, which are then fed into the agent for testing (see the upper part of \cref{fig:framework}).
The inverse solution is also rational: we can translate all English instructions into the target language and train the agent on the translated data, so it can be directly tested with target language instructions (see the lower part of \cref{fig:framework}).
The former agent is tested on translated instructions while the latter is trained on translated instructions. Both solutions suffer from translation errors and deviation from human instructions. 
But meanwhile, the former is trained on human-annotated English instructions (which we view as ``golden" data), and the latter is tested on ``golden" target language instructions. 
Motivated by this fact, we design a cross-lingual VLN framework that learns to benefit from both solutions. As shown in \cref{fig:framework}, we combine these two principles and introduce a cross-lingual language instructor (XLI), which learns to produce beliefs in the human instruction and its translation pair and to dynamically fuse the cross-lingual representations for better navigation.


After obtaining an efficient zero-shot agent, we investigate the question, \emph{can our cross-lingual VLN framework improve source-to-target knowledge transfer if given a certain amount of data for the target language?} 
We conduct extensive experiments to show that the cross-lingual language instructor lay an effective foundation for solving the circumstances that the agent has access to the source language and (partial) target language instructions for training. 
To validate our methods, we introduce a cross-lingual VLN dataset by collecting complimentary Chinese instructions for the English instructions in the Room-to-Room dataset~\citep{anderson2018vision}. Overall, our contributions are three-fold:
\begin{itemize}
    \item We collect the first cross-lingual VLN dataset to facilitate navigation agents towards accomplishing instructions of various languages such as English and Chinese.
    \item We introduce the task of cross-lingual vision-language navigation and propose a principled cross-lingual learning framework with a pretrained cross-lingual transformer and a cross-lingual language instructor.
    \item We demonstrate the efficiency of our model for cross-lingual knowledge transfer under two challenging settings, zero-shot learning where no target language data is available, and transfer learning where a certain amount of such data is given. 
\end{itemize}

\section{Problem Formulation}
The cross-lingual vision-language navigation task is defined as follows: we consider an embodied agent that learns to follow natural language instructions and navigate from a starting pose to a goal location in photo-realistic 3D indoor environments. 
Formally, given an environment $\mathcal{E}$, an initial pose $p_1=(v_1,\phi_1,\theta_1)$ (spatial position, heading, elevation angles) and natural language instructions $x_{1:N}$, the agent takes a sequence of actions $a_{1:T}$ to finally reach the goal $G$.
At each time step $t$, the agent at pose $p_t$ receives a new observation $\mathcal{I}_t=\mathcal{E}(p_t)$, which is a raw RGB image pictured by the mounted camera. Then it takes an action $a_{t}$ and leads to a new pose $p_{t+1} = (v_{t+1}, \phi_{t+1}, \theta_{t+1})$. 
After taking a sequence of actions, the agent stops when a \textit{stop} action is taken.

A cross-lingual VLN agent learns to understand different languages and navigate to the goal. 
Without loss of generality, we consider a bilingual situation coined as cross-lingual VLN.
To support the task, we built the cross-lingual VLN dataset $\mathcal{D}$, which includes human instructions in two different languages.
Specifically, $\mathcal{D}=\{(\mathcal{E}_i, p_{i,1}, x^\mathcal{S}_{i,1:N}, x^\mathcal{T}_{i,1:N'}, G_i)\}_{i=1}^{|\mathcal{D}|}$, where $\mathcal{S}$ and $\mathcal{T}$ indicate source and target language domains.
The footnote $i$ is eliminated in future discussions for simplicity.
Domain $\mathcal{S}$ contains instructions in the source language covering the full VLN dataset (including training and testing splits),
while domain $\mathcal{T}$ consists of a fully annotated testing set and a training set in the target language that covers a varying percentage $\epsilon$ of trajectories of the training set in $\mathcal{D}$ ($\epsilon$ may vary from 0\% to 100\%).
The agent is allowed to leverage both source and target language training sets and expected to perform navigation given an instruction from either source or target language testing sets. 

In this study, we first focus on a more challenging setting where no human-annotated target language data is available for training ($\epsilon = 0\%$), i.e., with only access to the source language training set, the agent is required to follow a target language instruction $x^\mathcal{T}_{1:N'}$ and navigate to the destination.
Then we investigate the agent's transferring ability by gradually increasing the percentage of human-annotated target language instructions for training ($\epsilon = 0\%, 10\%, ..., 100\%$). 

\begin{figure}
 \centering
 \begin{subfigure}[b]{0.24\textwidth}
     \centering
     \includegraphics[width=\textwidth]{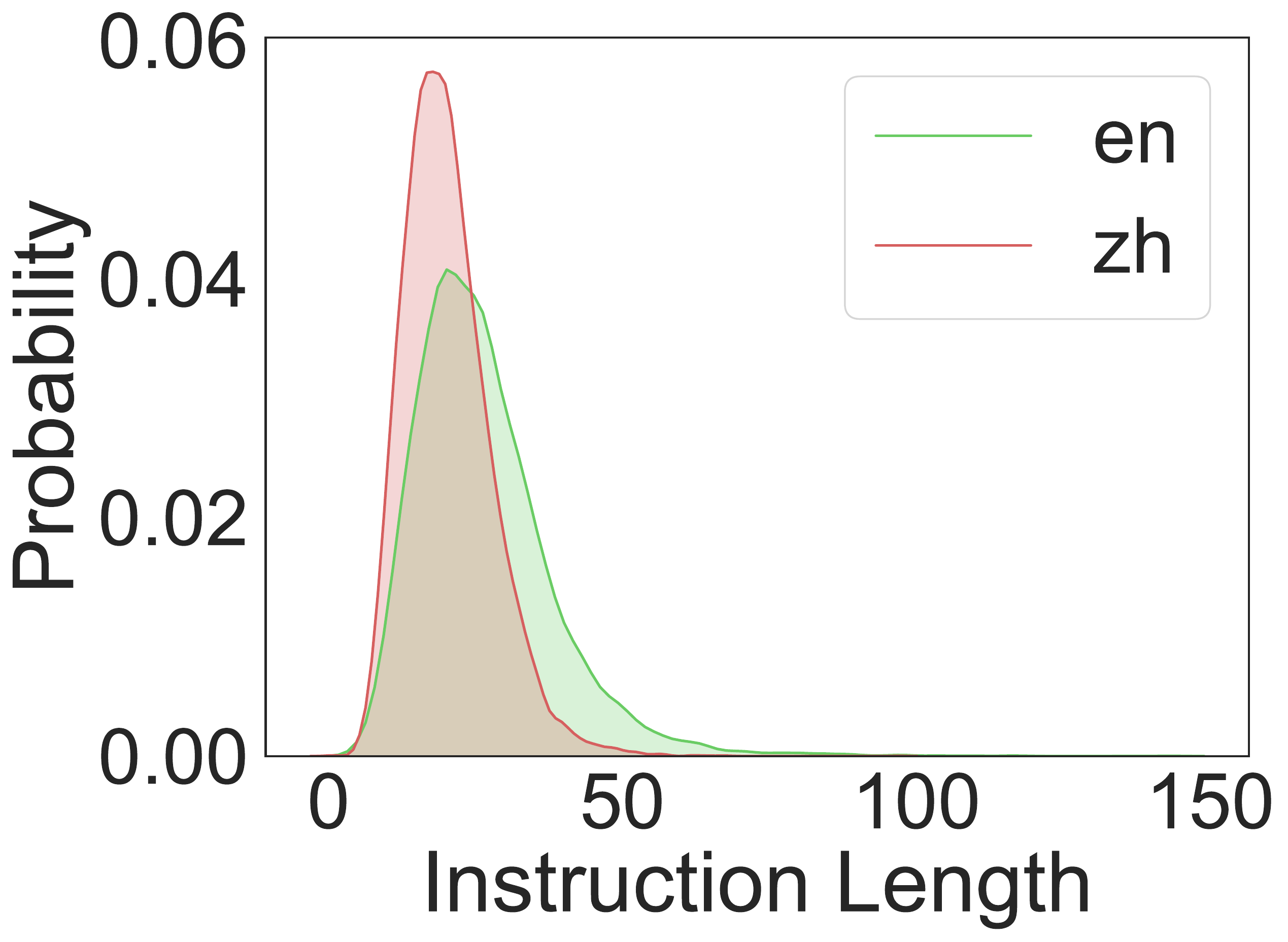}
     \caption{}
     \label{fig:ild}
 \end{subfigure}
 \begin{subfigure}[b]{0.229\textwidth}
     \centering
     \includegraphics[width=\textwidth]{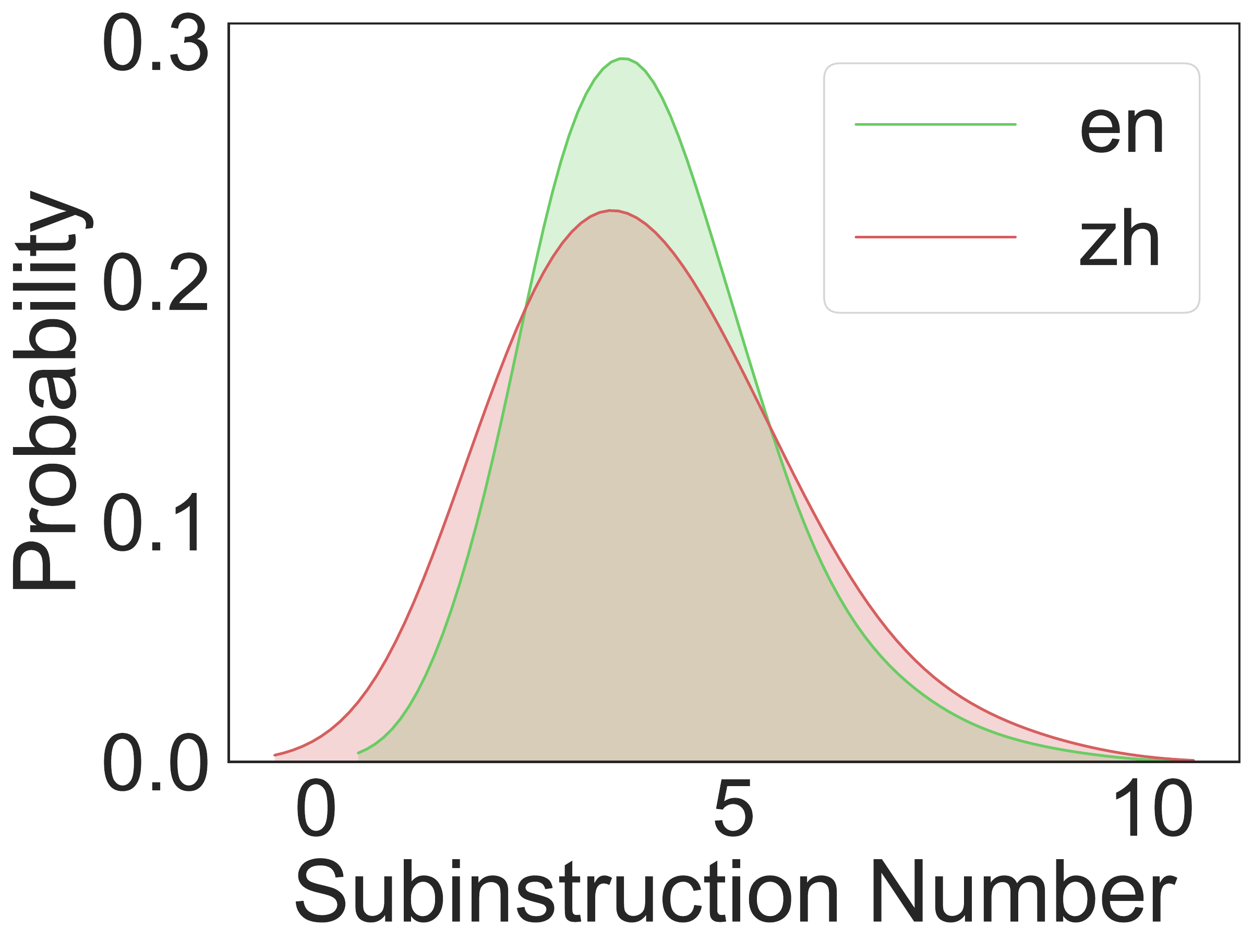}
     \caption{}
     \label{fig:sub-instr}
 \end{subfigure}
 
 \begin{subfigure}[b]{0.23\textwidth}
     \centering
     \includegraphics[width=\textwidth]{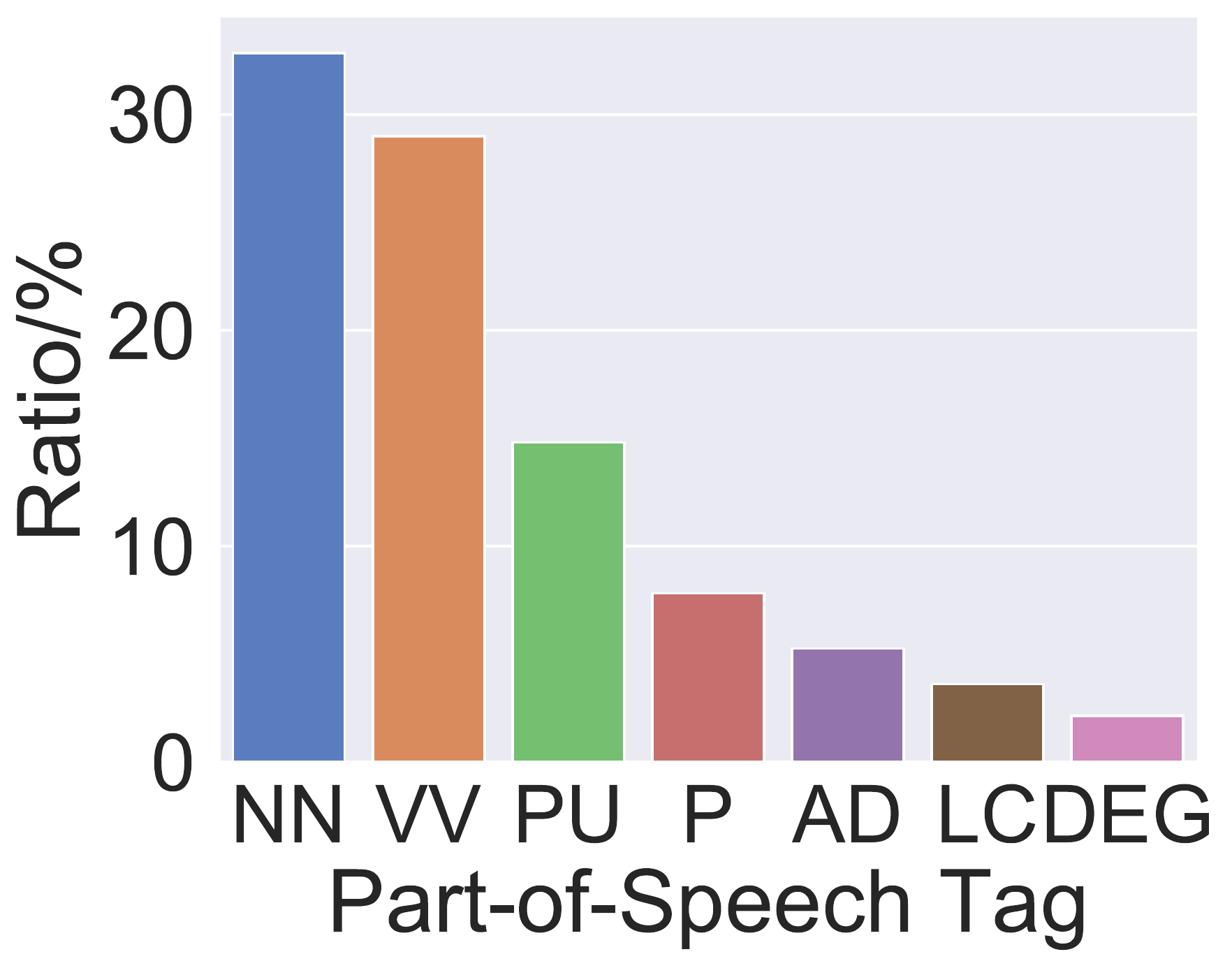}
     \caption{}
     \label{fig:zh-pos}
 \end{subfigure}
 \begin{subfigure}[b]{0.23\textwidth}
     \centering
     \includegraphics[width=\textwidth]{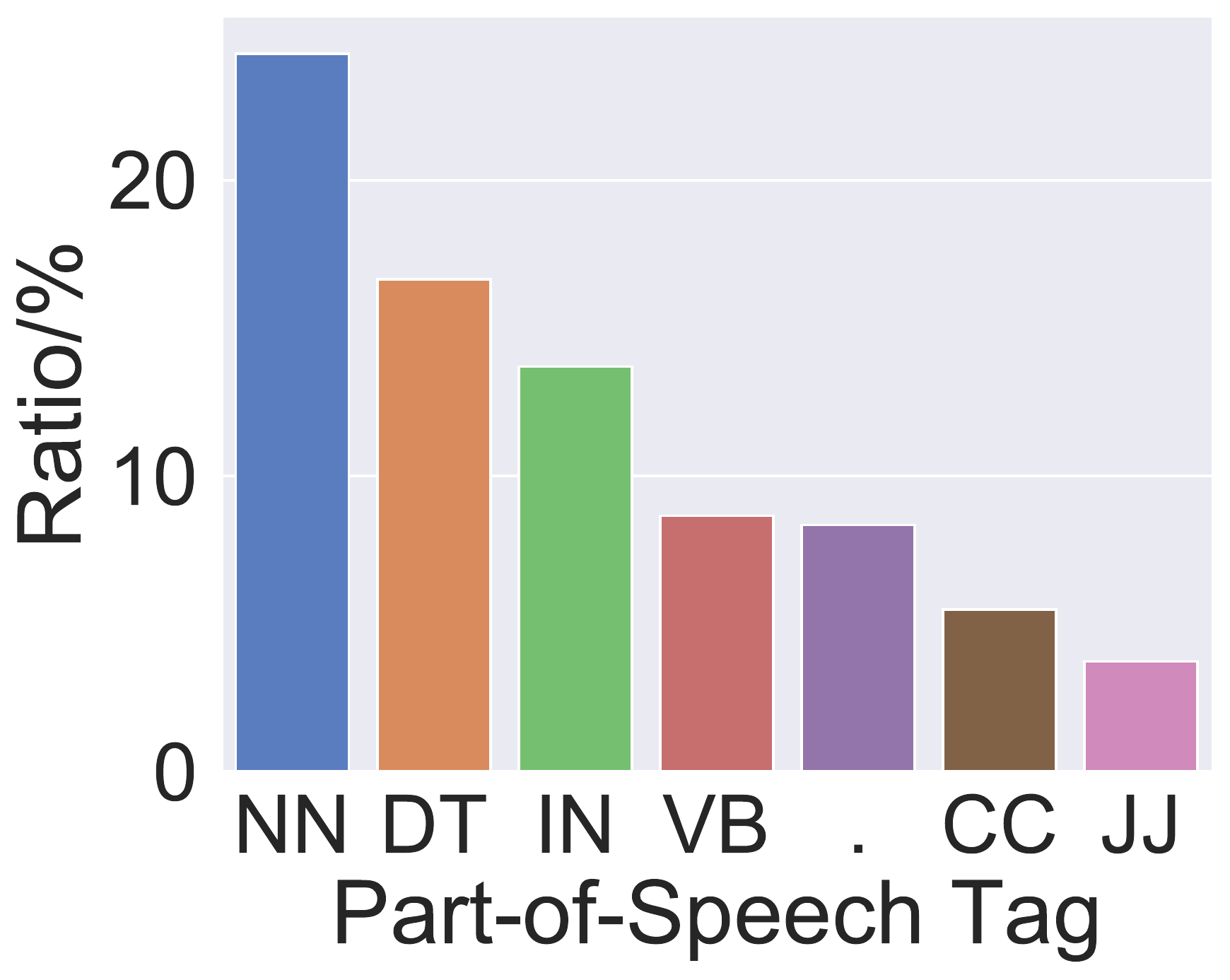}
     \caption{}
     \label{fig:en-pos}
 \end{subfigure}
 \caption{Analysis of XL-R2R English and Chinese corpora. (a) Distribution of instruction lengths. (b) Distribution of sub-instructions per instruction. (c) and (d) are distributions of part-of-speech tags for Chinese and English instructions.}
 \label{fig:en-zh}
\end{figure}

\section{XL-R2R Dataset}
We build a cross-lingual Room-to-Room (XL-R2R) dataset, 
the first cross-lingual dataset for the vision-language navigation task. 
It includes $4,675$ trajectories for the \emph{training} set, $340$ for \emph{validation seen}, and $783$ for \emph{validation unseen}, preserving the same split as the R2R dataset\footnote{Note that the original testing set of R2R is unavailable because the testing trajectories are held for challenge use.}.
Each trajectory is described with 3 English and 3 Chinese instructions independently annotated by different workers.

\subsection{Data Collection}
We keep the English instructions of the R2R dataset and collect Mandarin Chinese instructions via a public Chinese crowdsourcing platform.
The Chinese instructions are annotated by native speakers through an interactive 3D WebGL environment, following guidance by~\citet{anderson2018vision}. 
More details can be found in the Appendix.

\begin{figure}
 \centering
 \begin{subfigure}[h]{0.229\textwidth}
     \centering
     \includegraphics[width=\textwidth]{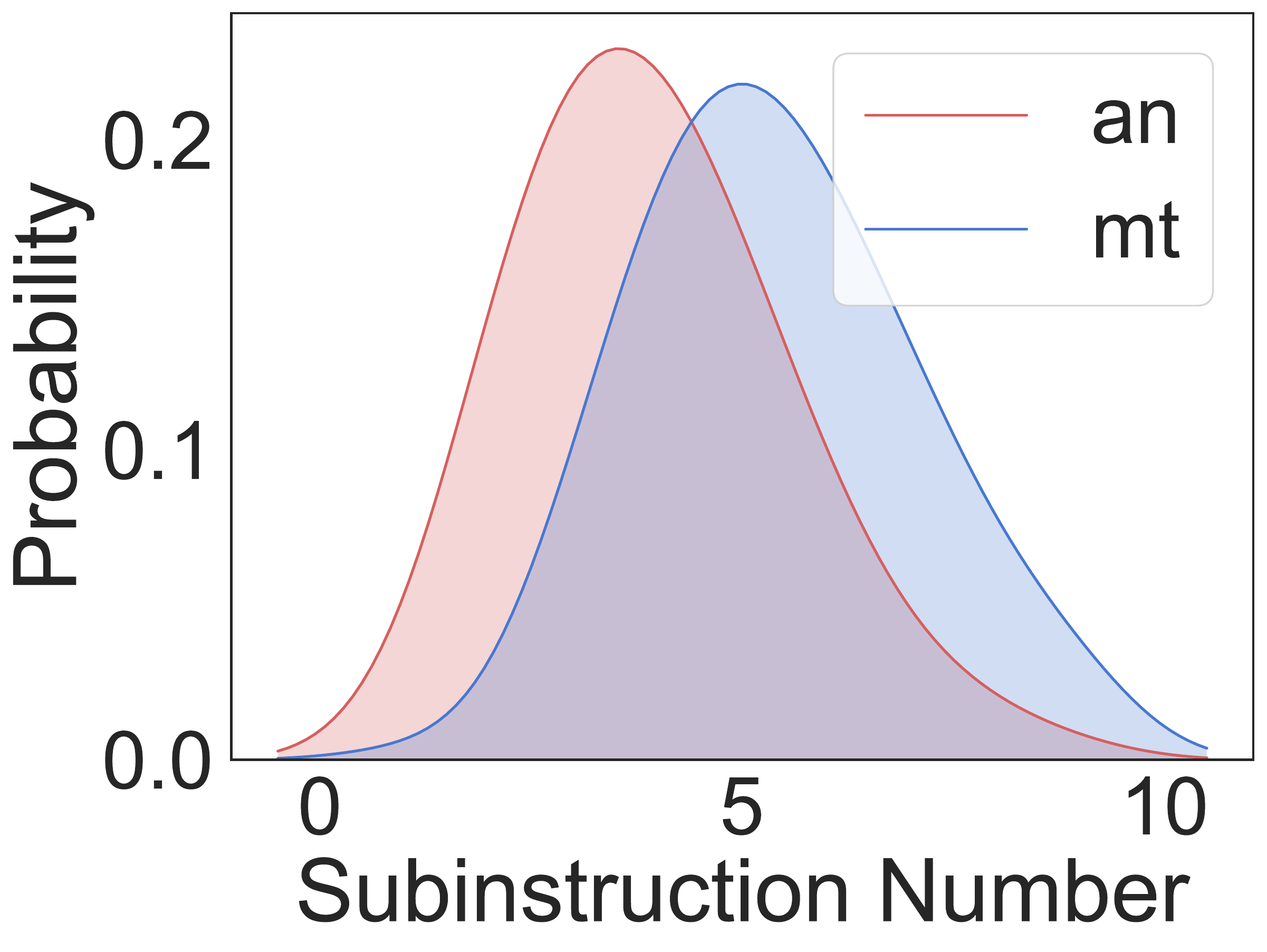}
     \caption{}
     \label{fig:sub-instr-anmt}
 \end{subfigure}
 \hfill
 \begin{subfigure}[h]{0.23\textwidth}
     \centering
     \includegraphics[width=\textwidth]{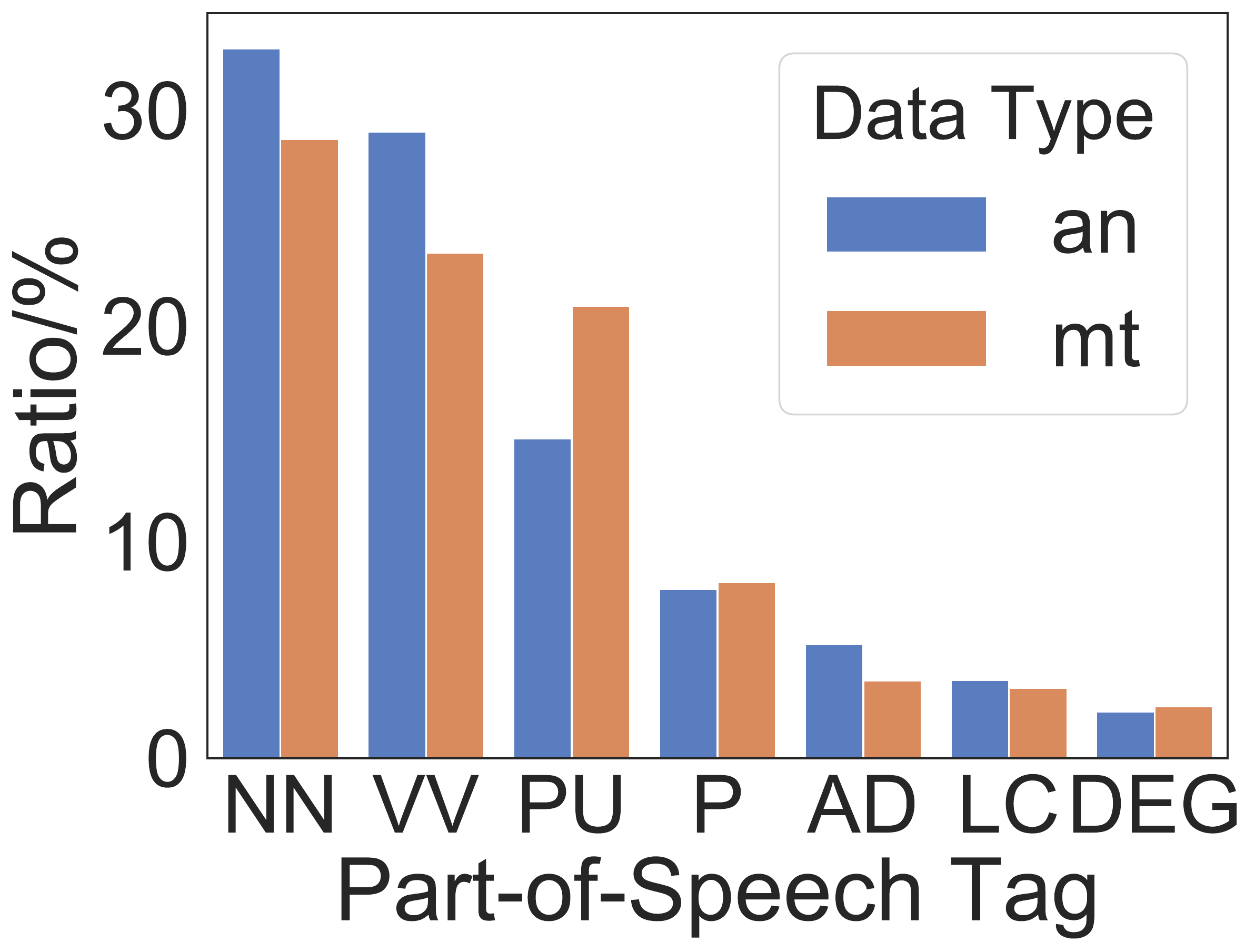}
     \caption{}
     \label{fig:zh-pos-anmt}
 \end{subfigure}
 \hfill
 \caption{Statistics of human annotated and machine translated data. 
 (a) is sub-instruction number per instruction distribution. (b) is top 7 part-of-speech tag distribution of annotated and machine translated instructions.}
 \label{fig:zh-an-mt}
\end{figure}

\subsection{Data Analysis}
\label{sec:data}
\paragraph{Chinese Annotations vs. English Annotations}
XL-R2R includes 17,394 instructions in both English and Chinese, annotated on 5,798 trajectories in total.
In \cref{fig:en-zh}, we compare statistics of the English and Chinese corpora. Note that we segment Chinese words using the Jieba\footnote{\url{https://github.com/fxsjy/jieba}} toolkit.
Removing words with less than 5 frequency, we obtain an English vocabulary of 1,583 words and a Chinese vocabulary of 1,134 words.
First, as shown in \cref{fig:ild}, Chinese instructions are shorter than English ones on average.
Second, the instructions usually consist of several sub-instructions separated by punctuation, and we can observe that the numbers of sub-instructions per instruction distribute similar across languages (\cref{fig:sub-instr}).
Furthermore, \cref{fig:zh-pos} and \cref{fig:en-pos} show that nouns and verbs, which often refer to landmarks and actions, are used more frequently in Chinese instructions (32.9\% and 29.0\%) than in English ones (24.3\% and 13.7\%)\footnote{The POS tags are obtained via Stanford Part-Of-Speech Tagger~\citep{toutanova2003feature-rich}.}.

  \begin{figure*}[t]
  \centering
  \includegraphics[width=.95\linewidth]{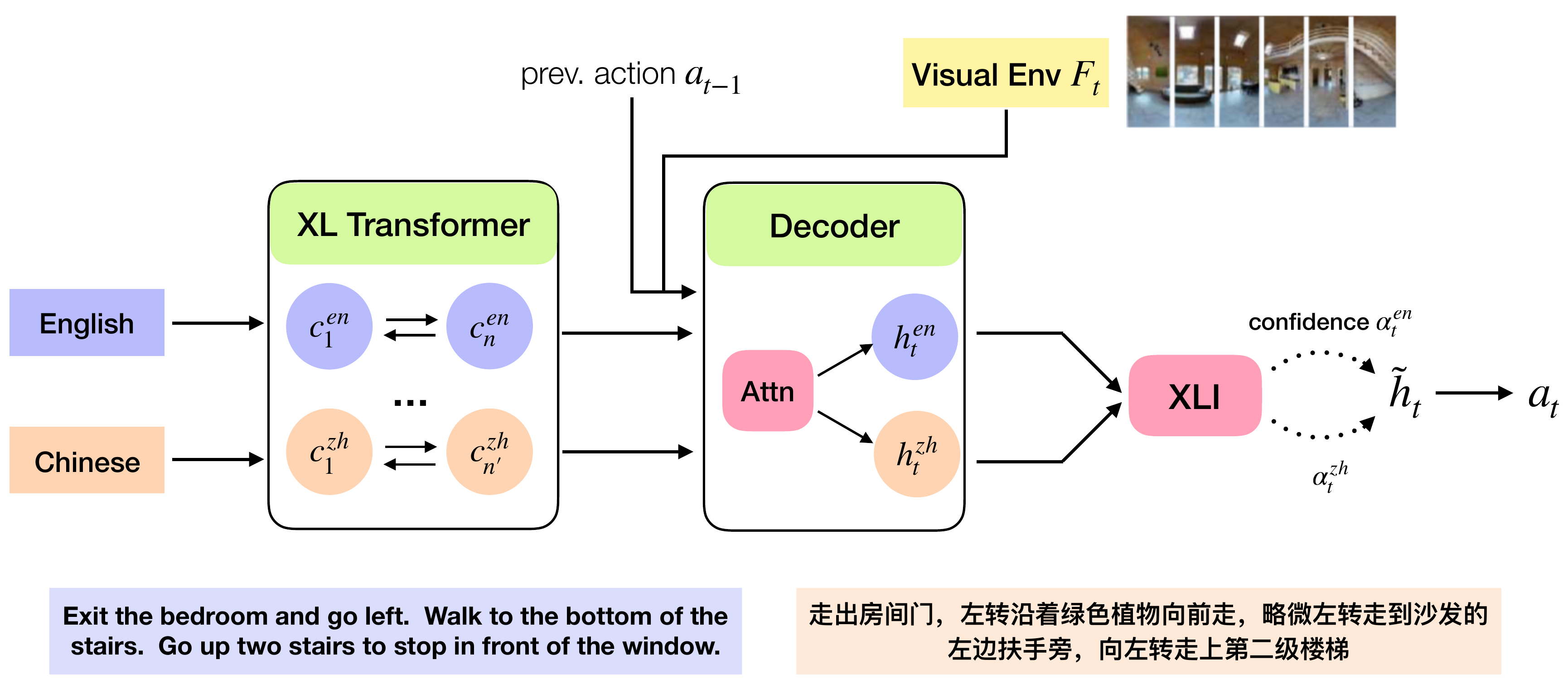}
  \caption{Illustration of the proposed cross-lingual VLN framework.}
  \label{fig:model}
  \end{figure*}

\paragraph{Chinese Annotations vs. Machine Translations}
In \cref{fig:zh-an-mt}, We compare the statistics of the Chinese annotated dataset with a machine-translated one.
The annotated instructions are more likely to contain fewer sentences per instruction.
Besides, nouns and verbs, which usually represent landmarks and actions in VLN task, are more frequent in annotated instructions than machine-translated ones, which shares the same trend as comparing Chinese and English annotations. 
The above analysis shows that annotations and machine translations have different data distributions, hence directly deploying a model trained with machine translations will lead to performance decrease.

\section{Method}
\label{sec:method}
We present a general cross-lingual VLN framework in \cref{fig:model}. It is composed of three modules: a pretrained cross-lingual transformer as the encoder, a decoder with panoramic action space and a cross-lingual language instructor (XLI).
Particularly, as shown in \cref{fig:model}, both English and Chinese instructions are encoded by a cross-lingual transfomer. 
Then the shared decoder takes the encoded contextual embeddings $c^\mathcal{I}_{1:N}$ from each language, the previous action $a_{t-1}$, and the local visual feature $\mathcal{F}_t$ as input, and produces hidden states $h_t^{en}$ for English and $h_t^{zh}$ for Chinese.  
The language instructor learns to assign probabilities to $h_t^{en}$ and $h_t^{zh}$, and then makes final predictions with the dynamically fused cross-lingual representation $\tilde{h}_t$. 

\paragraph{Cross-Lingual Transformer}
Motivated by recent advances in pretraining multilingual language models at scale, we leverage a pretrained cross-lingual transformer to enable cross-lingual zero-shot and transfer learning.
We employ the \textit{XLM-R} architecture in \citet{conneau2019unsupervised} to encode both languages.
It is a transformer model trained with multilingual masked language models (MLM) objective on 100 languages, which could address the data sparsity issue for our task with pretrained multilingual knowledge.

Receiving a pair of natural language instruction $x^\mathcal{I}_{1:N}, \mathcal{I}\in\{\mathcal{S}, \mathcal{T}\}$, the cross-lingual transformer encodes the sentence to obtain contextual word representations $c^\mathcal{I}_{1:N}$, along with the pooled hidden state $h^\mathcal{I}_{enc}$ at the first token of the instruction.

\paragraph{Panoramic Decoder}
At each time step, the agent perceives a 360-degree panoramic view of its surrounding scene from its current location with image feature $\mathcal{F}_t$, discretized into 36 view angles. Each view angle is represented by an encoding vector $v_i$.
The attended feature representation is computed with previous memory vector $s^\mathcal{I}_{t}$ :
\begin{align}
    e_{t,i} = \text{Attn}_{vis} (\mathcal{F}_t, s^\mathcal{I}_{t}) \\
    \mathcal{F}_{t,att} = \sum_{i} e_{t,i} \mathcal{F}_{t,i}
\end{align}

The decoder LSTM is initialized with $h^\mathcal{I}_{enc}$. It takes the concatenation of current attended image feature $\mathcal{F}_{t,att}$ and previous action embedding $a_{t-1}$ as input, and updates the hidden state from $s^\mathcal{I}_{t-1}$ to $s^\mathcal{I}_t$ aware of the historical trajectory:
\begin{equation}
    s^\mathcal{I}_{t} = \text{LSTM}_{\text{dec}}(s^\mathcal{I}_{t-1}, [\mathcal{F}_{t,att}, a_{t-1}])
    \label{eq:1}
\end{equation}
An attention mechanism is used to compute a weighted context representation, grounded on the instruction $c^\mathcal{I}_{1:N}$ by the hidden state $s^\mathcal{I}_{t}$, then obtain final hidden representation $h^\mathcal{I}_{t}$ for each language: 
\begin{align}
    h^\mathcal{I}_{t} &= \tanh(W[\tilde{c}^\mathcal{I}_{t}, s^\mathcal{I}_{t}]) \\
    \tilde{c}^\mathcal{I}_{t} &= \text{Attn}(c^\mathcal{I}_{1:N}, s^\mathcal{I}_{t})
\end{align}

\paragraph{Cross-lingual Language Instructor}
To bridge the gap between source and target languages, we leverage a production-level machine translation (MT) system
to translate the source language in the training data into the target language. 
During testing, the MT system will translate the target language instruction into the source language. 
The MT data serves as augmented data for zero-shot or low-resource settings as well as associates two different human languages in general. 
So we take two instructions (the human language instruction and its MT pair) as input for both training and testing. 
But we observed that these two instructions often generate different predictions, although one is the direct translation of the other. 
At each time step, when the agent observed the local visual environment, with two languages leading to different next positions, it remains a challenge which language representation to trust more.

Therefore, we propose a cross-lingual language instructor that learns to make the judgment.
At each time step, we let the language instructor decide which language representation we should have more faith in, i.e., ``learning to trust''.
The language instructor is a softmax layer 
which takes the concatenation of two hidden states $h^\mathcal{S}_{t}$ and $h^{\mathcal{T}}_{t}$ as input, and produces a probability $\alpha_t$ representing the belief of the source language representation. 
The final hidden vector used for predicting actions is defined as a mixture of the representations in two languages: 
\begin{equation}
\tilde{h}_{t} = \alpha_{t} h^\mathcal{S}_{t} + (1-\alpha_{t}) h^\mathcal{T}_{t}
\end{equation}3
Finally, the predicted action distribution for the next time step is computed as: 
\begin{equation}
    P(a_{t}|a_{1:t-1}, \mathcal{F}_{1:t}, x^\mathcal{S}_{1:N}, x^\mathcal{T}_{1:N})=\text{softmax}(\tilde{h}_{t})
\end{equation}

Thus the training objective is defined as the cross-entropy between the true actions and the predictive ones:
\begin{equation}
    \mathcal{L}_{XLI}=-\sum_t {\log P(a_{t}|a_{1:t-1}, \mathcal{F}_{1:t}, x^\mathcal{S}_{1:N}, x^\mathcal{T}_{1:N})}
\end{equation}

\section{Experiments}

\subsection{Experimental Setup}

\paragraph{Evaluation Metrics.}
The following evaluation metrics are reported: 
(1) Path length (PL), which measures the total length of predicted paths;
(2) Navigation Error (NE), mean of the shortest path distance in meters between the agent’s final location and the goal location;
(3) Success Rate (SR), the percentage of final positions less than 3m away from the goal location; 
(4) Oracle Success Rate (OSR), the success rate if the agent can stop at the closest point to the goal along its trajectory;
(5) Success rate weighted by (normalized inverse) Path Length (SPL)~\citep{anderson2018evaluation}, which trades-off Success Rate against trajectory length;
(6) Coverage weighted by Length Score (CLS)~\citep{jain2019stay}, which measures the fidelity to the described path and is complementary to goal-oriented metrics.

\paragraph{Implementation Details.}
We follow the same preprocessing procedure as in previous work~\citep{fried2018speaker}. 
A ResNet-152 model ~\citep{he2016deep} pretrained on ImageNet is used to extract image features, which are $2048$-d vectors. 
Instructions are clipped with a maximum length of $80$.
We use a XLM-RoBERTa base model ~\citep{conneau2019unsupervised} pretrained on 100 languages as the encoder.
The hidden sizes for encoder and decoder LSTM are $768$ and $512$ respectively. 
The dropout ratio is $0.5$. 
Each episode consists of no more than 10 actions.

The network is optimized via the ADAM optimizer~\citep{kingma2014adam} with initial learning rates of $1e-5$ on the pretrained encoder and $1e-4$ on the decoder, a weight decay of $0.0005$, and a batch size of $100$.
We train each model for $10,000$ iterations and evaluate it every $100$ iterations. We report the iteration with the highest SPL on validation unseen set. 

\begin{table*}[hbtp]
\small
\centering
\setlength{\tabcolsep}{5pt}
\begin{tabular}{c ccccc ccccc}
\toprule
 & \multicolumn{5}{c}{\textbf{Validation Seen}} & \multicolumn{5}{c}{\textbf{Validation Unseen}}\\
\cmidrule(lr){2-6} \cmidrule(lr){7-11}
Model & NE $\downarrow$ & OSR $\uparrow$ & SR $\uparrow$ & SPL $\uparrow$ & CLS $\uparrow$ & NE $\downarrow$ & OSR $\uparrow$ & SR $\uparrow$ & SPL $\uparrow$ & CLS $\uparrow$ \\
\toprule
train w/ MT & 5.58 & 54.7 & 41.9 {\scriptsize$\pm 6.0$} & 35.8 {\scriptsize $\pm 5.5$} & 53.5 {\scriptsize$\pm 3.7$} & 6.99 & 39.2 & 28.9 {\scriptsize$\pm 2.8$} & 21.5 {\scriptsize$\pm 2.7$} & 36.8 {\scriptsize$\pm 2.9$}\\
test w/ MT & 5.53 & 57.1 & 43.5 {\scriptsize$\pm 6.4$} & 37.1 {\scriptsize$\pm 6.3$} & 54.3 {\scriptsize$\pm 5.1$} & 7.32 & 39.3 & 27.9 {\scriptsize$\pm 4.7$} & 20.4 {\scriptsize$\pm 4.5$} & 35.5 {\scriptsize$\pm 4.1$} \\
\midrule
XLI & \textbf{5.12} & \textbf{59.0} & \textbf{48.4} {\scriptsize$\pm 0.8$} & \textbf{41.9} {\scriptsize$\pm 1.2$} & \textbf{57.7} {\scriptsize$\pm 1.8$} & \textbf{6.87} & \textbf{42.3} & \textbf{30.9} {\scriptsize$\pm 0.5$} & \textbf{23.4} {\scriptsize$\pm 0.8$} & \textbf{38.3} {\scriptsize$\pm 1.0$} \\
\midrule 
train w/ AN & 4.83 & 59.9 & 50.1 {\scriptsize$\pm 0.6$} & 43.2 {\scriptsize$\pm 0.1$}   & 58.1 {\scriptsize$\pm 0.3$} & 6.91 & 41.1 & 31.4 {\scriptsize$\pm 0.0$} & 23.9 {\scriptsize$\pm 0.2$} & 38.4 {\scriptsize$\pm 0.3$}\\
\bottomrule
\end{tabular}
\caption{Zero-shot learning results. Reported results are averages of 3 individual runs and shown with ($mean\pm std$). 
\textit{train w/ MT} denotes the model trained with Chinese MT data. 
\textit{test w/ MT} denotes the model trained with English annotations and test with English MT data translated from Chinese. 
\textit{XLI} is the framework presented in \cref{fig:framework} that aggregates two learning schemes with a cross-lingual language instructor (\cref{sec:method}).
The first three models are all for zero-shot learning. The last one, \textit{train w/ AN}, is trained with 100\% human-annotated Chinese data.
All models except \textit{test w/ MT} are tested with human-annotated Chinese instructions.
}
\label{tab:zeroshot}
\end{table*}

\begin{figure*}[t]
    \centering
    \begin{subfigure}[b]{0.48\linewidth}
        \centering
        \includegraphics[width=\linewidth]{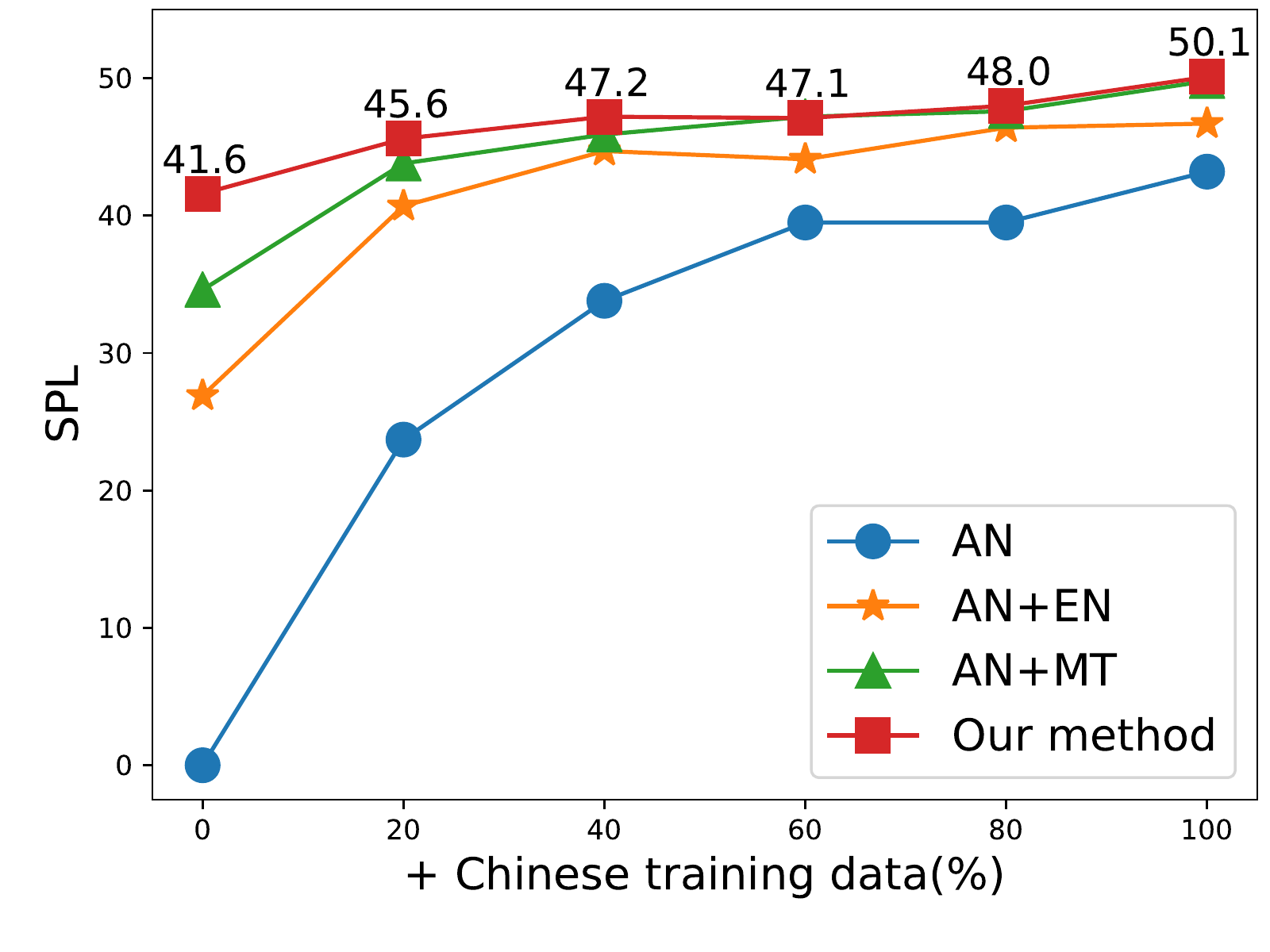}
        \caption{Results on Val Seen}\label{fig:transfer-seen}
    \end{subfigure}
    ~
    \begin{subfigure}[b]{0.48\linewidth}
        \centering
        \includegraphics[width=\linewidth]{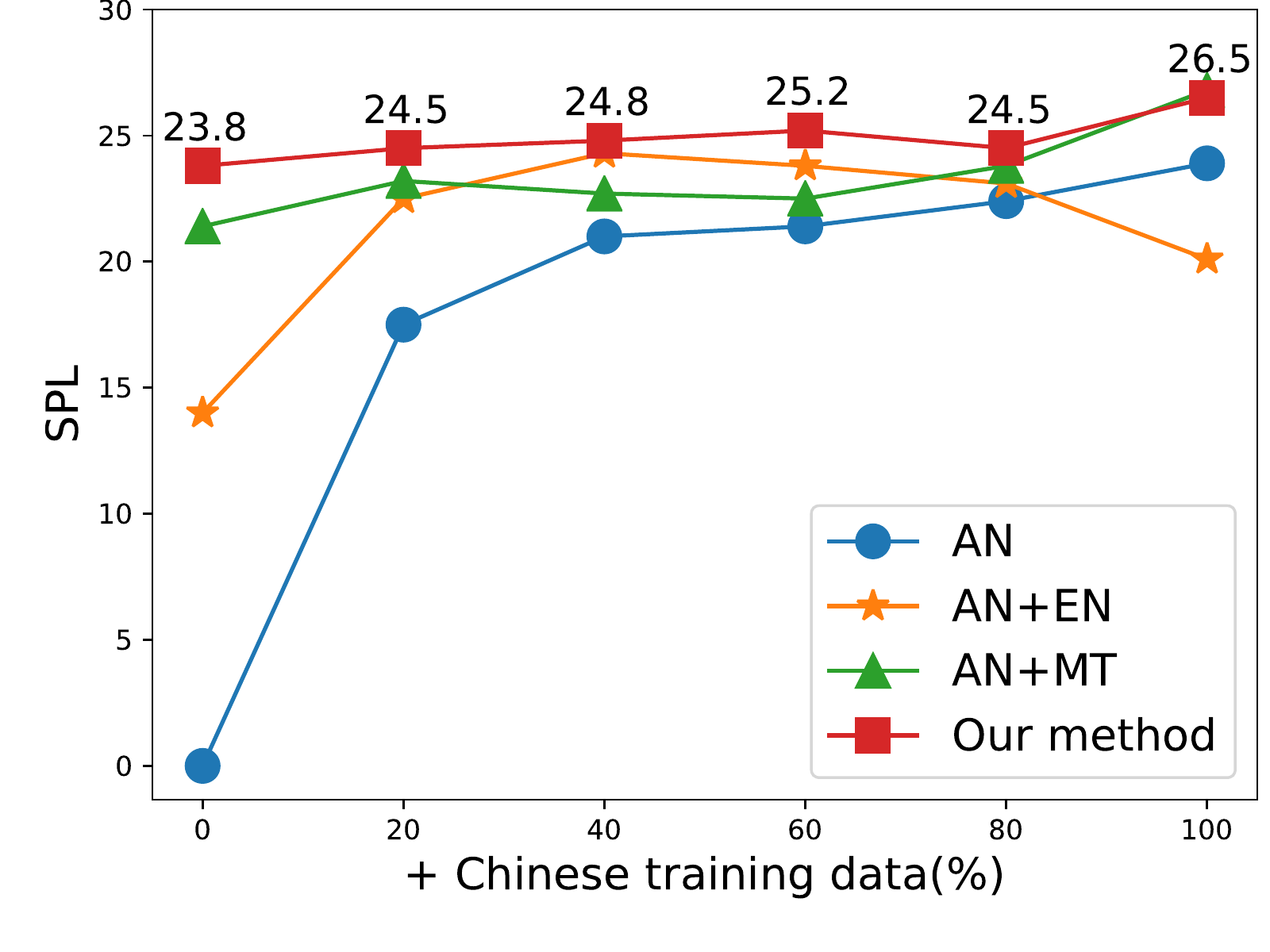}
        \caption{Results on Val Unseen}\label{fig:transfer-unseen}
    \end{subfigure}
\caption{We examine the influence of using different percentages (from 0\% to 100\%) of target language (Chinese) instructions to train agents. 
We compare the results on XLI with three baselines. \textit{AN}: only partial Chinese annotations (0\% to 100\%) are used for training. 
\textit{AN+EN}: partial Chinese annotations and 100\% English annotations. \textit{AN+MT}: 
partial Chinese annotations and 100\% Chinese translations of those English instructions. }
\label{fig:transfer}
\end{figure*}

\subsection{Zero-shot Learning}

We report results under the zero-shot setting in \cref{tab:zeroshot}, to show the effectiveness of our method. 
First, the difference between \textit{train w/ MT} and \textit{train w/ AN} validates our findings in \cref{sec:data}, that mismatched distributions between machine translations and human annotations will result in performance decrease, which indicates the insufficiency of solely using MT data for zero-shot learning. 
Second, the clear gap between \textit{train w/ MT} and \textit{XLI} proves that our language instructor can successfully aggregate cross-representations of both human-annotated and MT data.
Moreover, even though the agent does not have access to any annotated target language data, it achieves competitive results compared to \textit{train w/ AN} that is trained with 100\% annotated target language data.

\subsection{Transfer Learning}
To investigate the knowledge transfer effect from English to Chinese, we draw performance curves of utilizing varying percentages of Chinese annotations for training (see \cref{fig:transfer}). 
Particularly, the starting point is our zero-shot setting, where one has no access to human-annotated data of the target language (Chinese), and the endpoint is where one has 100\% training data of the target language. 

\cref{fig:transfer} demonstrates that the proposed approach provides consistent improvements over other methods in both seen and unseen environments. 
First, our method works for both low-resource and high-resource settings, and improves the transferring ability steadily as the size of Chinese annotations grows.
Besides, our method trained with 20\% Chinese annotations has already outperformed the results as the model trained with 100\% Chinese annotations. 
This demonstrates the sample efficiency of our cross-lingual VLN model and the potential of scaling it for more languages with only a small amount of annotated data required.  
Finally, one can also observe that training with both English and MT Chinese data helps learn useful encoding that is especially valuable when only limited Chinese training data is available. 

\begin{table*}
\small
\centering
\setlength{\tabcolsep}{4pt}
\begin{tabular}{cc cccc c cccc c}
\toprule
 \multirow{2}{*}{Encoder} & \multirow{2}{*}{\shortstack{Training\\ data}} & \multicolumn{5}{c}{\textbf{Validation Seen}} & \multicolumn{5}{c}{\textbf{Validation Unseen}}\\
\cmidrule(lr){3-7} \cmidrule(lr){8-12}
 &  & NE $\downarrow$ & OSR $\uparrow$ & SR $\uparrow$ & SPL $\uparrow$ &  $\Delta$ SPL & NE $\downarrow$ & OSR $\uparrow$ & SR $\uparrow$ & SPL $\uparrow$  &  $\Delta$ SPL \\
\toprule
\multirow{2}{*}{LSTM}& MT & 5.78 & 54.8 & 42.7 & 35.8 & \multirow{2}{*}{2.7}   & 7.56 & 34.8 & 25.1  & 18.8 & \multirow{2}{*}{1.3}\\
& AN & 5.18 & 60.0 & 47.2 & 38.5 &    & 7.06 & 40.4 & 28.8 & 20.1 \\

\midrule 
\multirow{2}{*}{M-BERT} & MT & 5.26 & 58.8 & 46.7 & 39.6 & \multirow{2}{*}{7.3}  & 6.92 & 42.5 & 29.8 & 21.4 &  \multirow{2}{*}{2.0}\\
 & AN & 4.68 & 62.9 & 53.6 & 46.9  & & 6.86 & 43.6 & 31.6 & 23.4\\

\midrule 
\multirow{2}{*}{XLM-R} & MT & 5.58 & 54.7 & 41.9  & 35.8 & \multirow{2}{*}{7.4}  & 6.99 & 39.2 & 28.9 & 21.5 &  \multirow{2}{*}{2.4}\\
 & AN & 4.83 & 59.9 & 50.1 & 43.2  & & 6.91 & 41.1 & 31.4 & 23.9\\

\bottomrule
\end{tabular}
\caption{Performance comparison for different encoders. \textit{LSTM} is to train the model with an LSTM encoder from scratch. \textit{M--BERT} is an uncased multi-lignual BERT-base model pretrained with masked language modeling and Next sentence prediction on 102 languages ~\citep{devlin2018bert}. \textit{XLM-R} is a cross-lingual RoBERTa-base model 
, which is the encoder we mainly used for this task. $\Delta$ SPL is the difference of two SPLs trained with machine translations and human annotations.}
\label{tab:encoders}
\end{table*}

\subsection{Encoder Variations}
To enable cross-lingual VLN, we examine the navigation performance with different types of encoders as the backbone. As shown in \cref{tab:encoders}, pretrained language models provide a better contextual representation of the instruction, hence leads to better navigation performance with the same decoder and training scheme.
Furthermore, \textit{M--BERT} and \textit{XLM-R} have similar performance for our task. 
Vision-language navigation could serve as a new benchmark for evaluating these language models, i.e., instead of common automatic metrics such as BLEU-4~\citep{papineni2002bleu} or ROUGE-L~\citep{lin-2004-rouge}, we can directly evaluate the performance of various language models with the navigation results.
Finally, the gaps ($\Delta SPL$) exist and are similar across all models when trained with different source data, i.e., machine translations and human annotations.

\subsection{Results on English Test Set}
We submitted the results to the VLN test server to evaluate the proposed approach on the unseen test set.
We treat English as the target language for both zero-shot learning and transfer learning with 100\% target training data. Results are presented in \cref{tab:test-en}.
For zero-shot learning, the agent has access to all human-annotated Chinese data but no English data during training. At test time, it is commanded to follow human-annotated English instructions. 
As shown in \cref{tab:test-en}, our method (XLI) improves by 6.3\% relatively over the model trained with MT data. 
For transfer learning, our method can efficiently transfer knowledge between Chinese and English data. The results here show similar trends as in the reported results on the Chinese validation set. (See \cref{tab:zeroshot} and \cref{fig:transfer}).

\begin{table}
\small
\centering
\setlength{\tabcolsep}{3pt}
\begin{tabular}{c cccc c}
\toprule
\multirow{2}{*}{Model} & \multicolumn{4}{c}{\textbf{Test (unseen)}} & \multirow{2}{*}{\shortstack{Access to target\\ training data}}\\
\cmidrule(lr){2-5}
 & NE $\downarrow$ & OSR $\uparrow$ & SR $\uparrow$ & SPL $\uparrow$ & \\
\toprule
train w/mt & 7.3 & 37.0 & 28.3 & 21.4 & \xmark\\
XLI (0\%) & \textbf{7.2} & \textbf{40.0} & \textbf{29.4} & \textbf{22.8}& \xmark \\
\midrule
train w/an & 7.0 & 39.7 & 30.6 & 24.1 & \cmark \\
XLI (100\%)  &  \textbf{6.7} & \textbf{42.8} & \textbf{33.0} & \textbf{25.2} & \cmark \\
\bottomrule
\end{tabular}
\caption{Results on the R2R English test set. The first two rows are for zero-shot learning, the last two rows are trained with access to 100\% target training data (i.e., annotated English instructions).}
\label{tab:test-en}
\end{table}

\subsection{Case Study}

\begin{figure*}[hbtp]
  \centering
  \includegraphics[width=.95\linewidth]{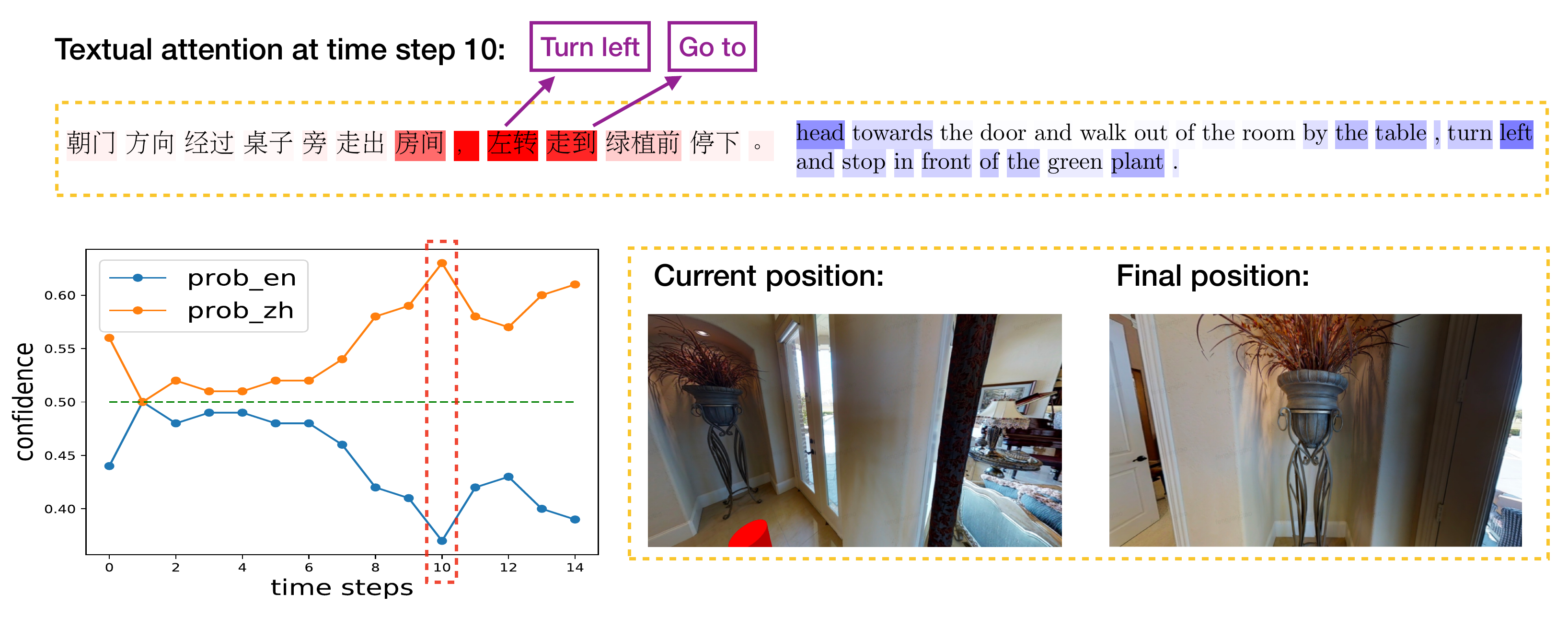}
  \caption{Case study. We choose a completed instruction from the validation set for illustration.
  }
  \label{fig:case}
\end{figure*}

For a more intuitive understanding of the language instructor, we visualize the confidences assigned to each language in \cref{fig:case}. In this case, the language instructor trusts more in the human-annotated Chinese instruction, which is of better quality. 
More specifically, at time step 10, when the language instructor has the highest faith in the Chinese instruction, we visualize the textual attention on both instructions at this time step. 
Evidently, the corresponding textual attention on the Chinese command makes more sense than that on the machine-translated English command. 
The agent is supposed to keep turning left and then move forward to the green plant. The attention on the Chinese instruction assigns 0.25 to  ``turn left'', and nearly zero weight to ``head towards the door'' which is already completed by previous actions.
The attention weights on English are more uniformly distributed and thus appear to be less accurate than that of Chinese.

\section{Related Work}
\paragraph{Vision and Language Grounding} 
Over the past years, deep learning approaches have boosted the performance of computer vision and natural language processing tasks \citep{chen2011learning,krizhevsky2012imagenet,sutskever2014sequence, he2016deep, vaswani2017attention}. 
A large body of benchmarks are proposed to facilitate the research, including image and video caption \citep{lin2014microsoft, krishna2017dense, xu2016msr}, VQA \citep{antol2015vqa, das2018embodied}, and visual dialog \citep{das2017visual}. 
These tasks require grounding on both visual and textual modalities, but mostly limited to a fixed visual input. 
Thus, we focus on the task of vision-language navigation (VLN) \citep{anderson2018vision}, where an agent needs to actively interact with the visual environment following language instructions. 

\paragraph{Vision-Language Navigation} 
Several approaches have been proposed for the VLN task on the R2R dataset. 
For example, \citet{wang2018look} presented a planned-ahead module combining model-free and model-based reinforcement learning methods, \citet{fried2018speaker} introduced a speaker, which can synthesize new instructions and implement pragmatic reasoning. 
Subsequent methods extend the speaker-follower model with reinforced cross-modal matching \citep{wang2019reinforced}, self-monitoring \citep{ma2019self}, back-translation \citep{tan2019learning} etc. 
Previous works mainly improve navigation performance by data augmentation or leveraging efficient searching methods.
In this paper, we address the task from a cross-lingual perspective, aiming at building an agent to execute instructions for different languages.

\paragraph{Cross-Lingual Language Understanding}
Learning cross-lingual representations is a crucial step to make natural language tasks scalable to all the world's languages. 
Cross-lingual studies on typical NLP tasks have achieved success, such as part-of-speech tagging \citep{zhang2016ten, kim2017cross}, sentiment classification \citep{zhou2016cross, chen2018adversarial}, named entity recognition \citep{pan2017cross, ni2017weakly} and vision-language tasks\citep{kim2019mule, miyazaki2016cross, wang2019vatex}.
These studies successfully disentangle the linguistic knowledge into language-common and language-specific parts
with individual modules.

Recently, bidirectional transformers ~\citep{devlin2018bert, yang2019xlnet} pretrained on large-scale corpus data has drawn significant attention in the community. Its cross-lingual variants such as XLM and XLM-RoBERTa \citep{conneau2019cross,  conneau2019unsupervised} showed superior performance on a wide range of down-stream cross-lingual transfer tasks.
Our dataset and method address cross-lingual representation learning for the 
vision-language navigation task.
To our knowledge, we are the first to study cross-lingual learning in a dynamic visual environment, where the agent needs to interact with its surroundings and take a sequence of actions.

\section{Conclusion and Future Work}
In this paper, we introduce a new task, namely cross-lingual vision-language navigation, to study cross-lingual representation learning situated in the navigation task where cross-modal interaction with the real world is involved.
We collect a cross-lingual R2R dataset and conduct pivot studies towards solving this challenging but practical task. 
The proposed cross-lingual VLN framework shows its effectiveness in cross-lingual knowledge transfer.
There are still lots of promising future directions for this task and dataset, e.g., to incorporate recent advances in VLN and improve the model capacity. It would also be valuable to extend the dataset to support numerous different languages in addition to English and Chinese, as well as evaluate the cross-lingual performance of variant language models on this new benchmark.

\bibliography{anthology,eacl2021}
\bibliographystyle{acl_natbib}

\clearpage
\appendix
\onecolumn
\section*{Appendix}




\section{Chinese Data Collection}

\begin{figure}[htbp]
 \centering
 \includegraphics[width=\textwidth]{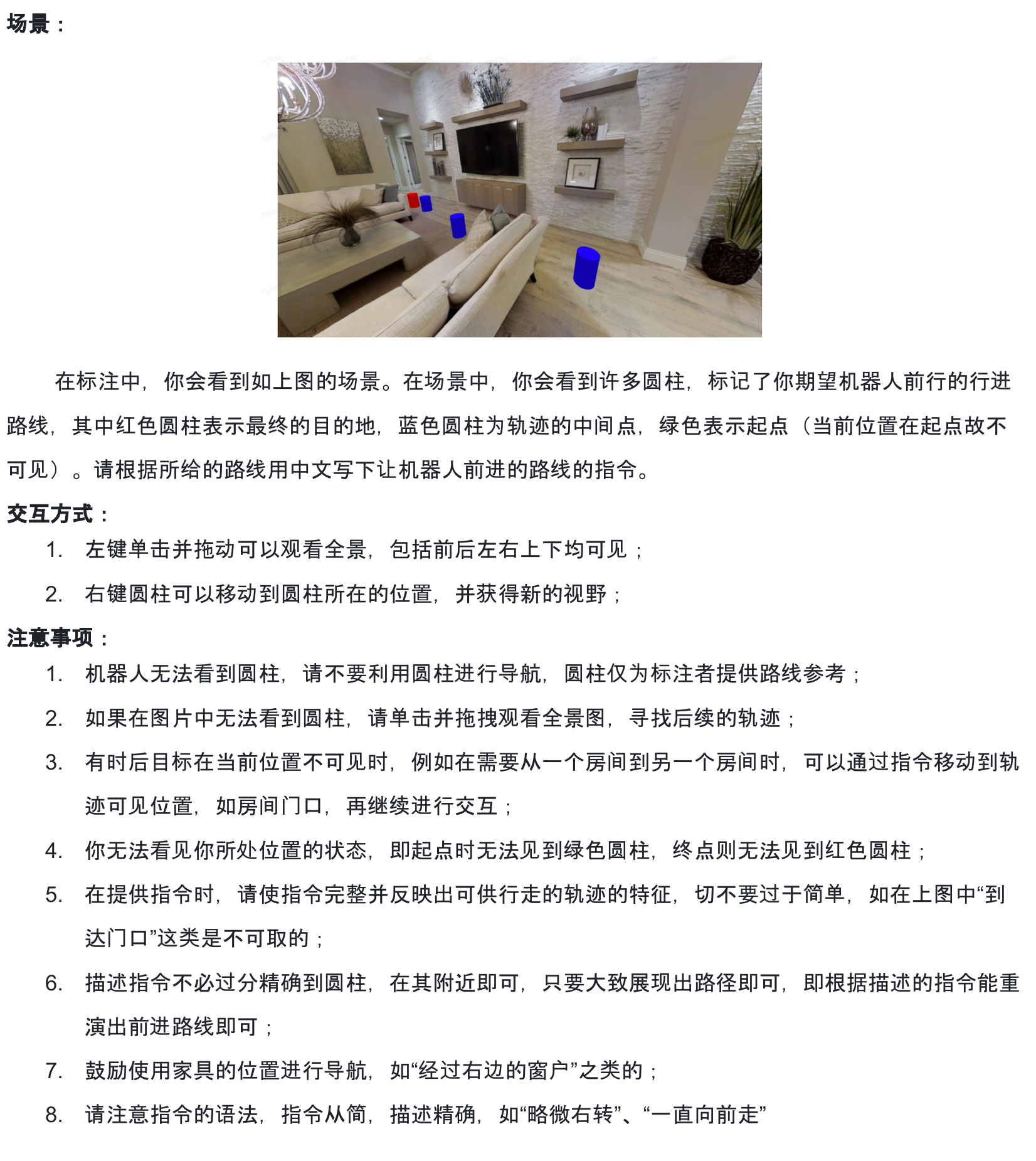}
 \caption{Instructions for Chinese data collection.}
 \label{fig:instrs}
\end{figure}

We paid 25 workers to do the data collection work via a public Chinese data collection platform, taking around 4 weeks to finish the task.
The workers are paid reasonably, with an estimated hourly rate higher than the local minimum wage.
Before starting annotation, we educated the workers in a face-to-face way with documented instructions to help them understand the task(see ~\cref{fig:instrs} ).

\begin{figure}[htbp]
 \centering
 \includegraphics[width=0.9\textwidth]{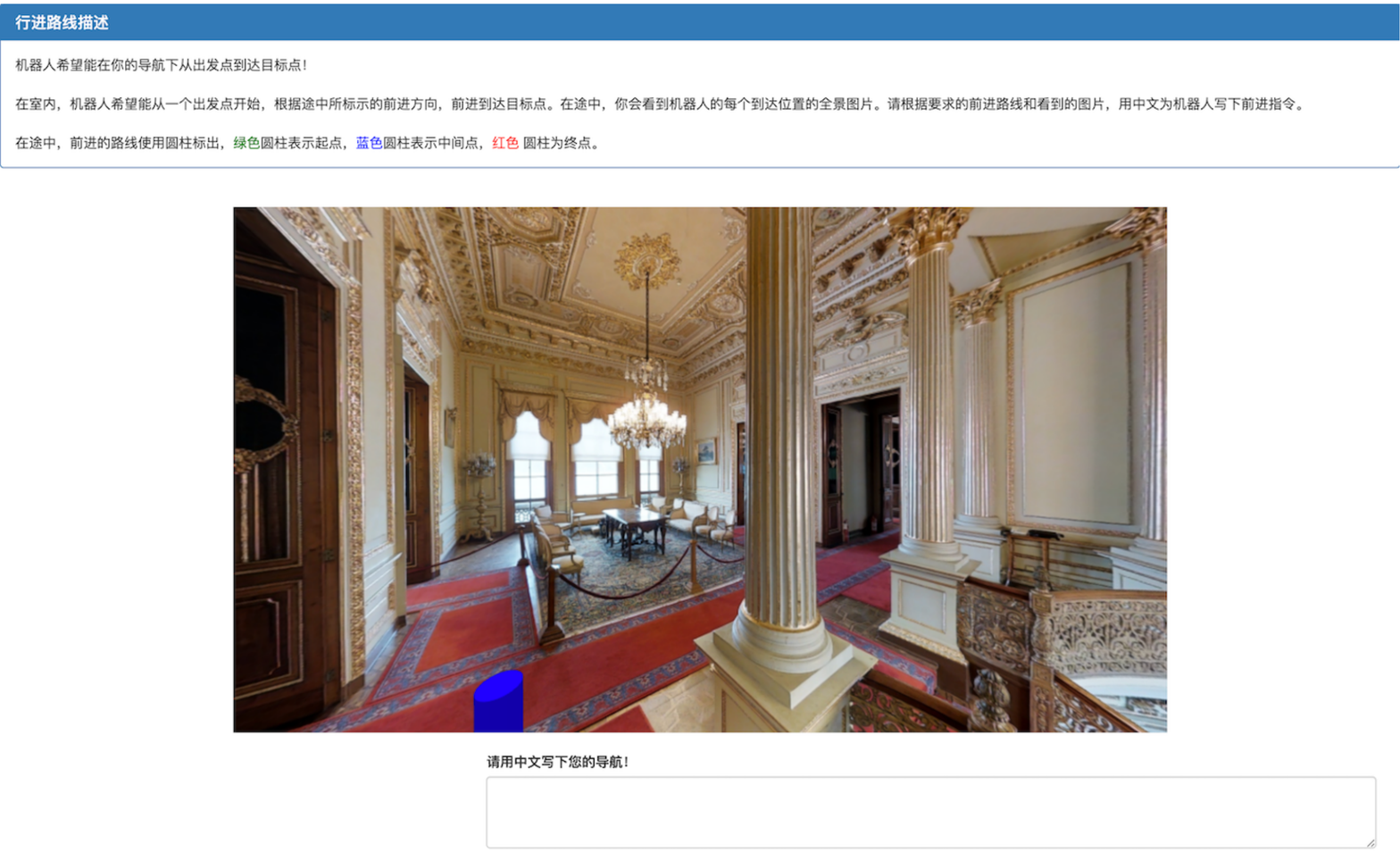}
 \caption{Interface for Chinese data collection.}
 \label{fig:interface}
\end{figure}

\end{document}